\pdfoutput=1

\documentclass[11pt]{article}

\usepackage{acl}

\usepackage{times}
\usepackage{latexsym}

\usepackage[T1]{fontenc}

\usepackage[utf8]{inputenc}

\usepackage{microtype}

\usepackage{graphicx}
\usepackage{listings}
\usepackage{amsmath}
\usepackage{amssymb,amsfonts}
\usepackage{algorithm}
\usepackage{algpseudocode}
\usepackage{hyperref}
\usepackage{caption}
\everymath{\displaystyle}

\title{Mapping Wordnets on the Fly with Permanent Sense Keys}
\author{Eric Kafe\\
  MegaDoc\\
  Charlottenlund, Denmark\\
  {\tt kafe@megadoc.net}
}
\date{}
\begin{document}
\maketitle
\addtolength{\itemsep}{0.5\baselineskip}
\begin{abstract}
Most of the major databases on the semantic web have links to
Princeton WordNet (PWN) synonym set (synset) identifiers, which differ for each PWN release,
and are thus incompatible between versions. On the other hand, both PWN and the more
recent Open English Wordnet (OEWN) provide permanent
word sense identifiers (the sense keys), which can solve this interoperability problem.

We present an algorithm that runs in linear time, to automatically derive a synset mapping
between any pair of Wordnet versions that use PWN sense keys. This allows to update old WordNet
links, and seamlessly interoperate with newer English Wordnet versions for which
no prior mapping exists.

By applying the proposed algorithm on the fly, at load time,
we combine the Open Multilingual Wordnet (OMW 1.4, which uses old PWN 3.0 identifiers)
with OEWN Edition 2021, and obtain almost perfect precision and recall.
We compare the results of our approach using respectively synset offsets, versus the
Collaborative InterLingual Index (CILI version 1.0) as synset identifiers, and find that
the synset offsets perform better than CILI 1.0 in all cases, except a few ties.
\end{abstract}

\section{Introduction}

All the available multilingual wordnets \citep{Bond14} and important knowledge bases
on the semantic web \citep{Babel, Sumo, Yago, Imagenet} were originally linked to different
versions of {\it Princeton WordNet} (PWN) \citep{WordNet}, using version-specific 
{\it synset offsets} \citep[Wndb]{wnm},
which differ between releases, so mappings are necessary for interoperation, and for updating
to a later English Wordnet versions.

Many of these resources have been remapped to Wordnet 3.0 or Wordnet 3.1,
using {\it offset to offset} mappings obtained by relaxation labelling \citep{daude-etal-2000-mapping},
{\it offset to ILI} (InterLingual Index) mappings
\citep{Voss02, vossen-etal-2016-toward, bond-etal-2016-cili},
{\it sensekey to sensekey} mappings \citep[Sensemap]{wnm},
and {\it offset to offset} mappings relying on sense key persistence \citep{kafe18}.
Contrary to synset offsets, the sensekeys persist across database versions \citep[Senseidx]{wnm},
and can thus support the derivation of mappings with high precision and recall.

PWN {\it sensekeys} \citep[Senseidx]{wnm} are composite database keys
representing one particular word sense.
They consist in the concatenation of the identifiers for the
corresponding  {\it lemma} and its {\it lexfile}, {\it lex\_id}, 
and eventually {\it head adjective} (see examples in sections 2.1, 3.2 and 4.2). 
Each PWN version includes an {\it index.sense} file, linking the sense keys to their
corresponding synset offsets.

However, the necessary mappings between synsets linked to different PWN versions
are not always available,
either because a resource is too new, or has too few users to justify the production of a mapping.
This causes potentially long delays for interoperability, which may remain
impossible as long as no relevant mapping exists. For example,
{\it Edition 2022} of the {\it Open English Wordnet} 
(OEWN\footnote{\url{https://github.com/globalwordnet/english-wordnet}}) 
\citep{mccrae-etal-2020-english} was released recently,
and the {\it wndb}\footnote{\url{https://github.com/x-englishwordnet/wndb}}
project has also published the same data in a PWN-compatible format (including the relevant
{\it index.sense}). These two variants of the OEWN 2022 Edition use different, mutually incompatible
synset offsets; no mapping exists for neither yet, and no known project currently aims
to produce such mappings.

On the other hand, OEWN has adopted PWN sensekeys as its main sense identifier,
so it is easy to extract a sense index from the database, and almost instantly
produce a sensekey-based mapping, since this only requires joining the 
{\it index.sense} of the relevant wordnet versions.
Therefore, we propose to carry out the mapping process on the fly, whenever loading
wordnets that are linked to different English Wordnet (PWN or OEWN) versions.

\section{Methods}

\subsection{Mapping Strategy}

Between two Wordnet versions, word senses can be either {\it added} or {\it removed},
and the same applies to synonym sets, in the case where all their elements are respectively
completely new or entirely deleted. In addition to that, synonym sets can also
be {\it split} and/or {\it merged}, when one or more of their elements are moved to another
(existing or new) synset.

For example, between versions 3.0 and 3.1 of PWN, {\it Pluto} was moved from the
{\it god of the underworld} in Greek mythology, to the synset with the names of the
corresponding Roman "god of the underworld":

\smallskip
\begin{center}
\resizebox{0.48\textwidth}{!}{
\begin{tabular}[]{lrrrr}
{\bf Sense Key}&{\bf PWN}$_{3.0}$&{\bf CILI}$_{3.0}$&{\bf CILI}$_{3.1}$&{\bf PWN}$_{3.1}$\\
\hline\noalign{\smallskip}
aides\%1:18:00::&09570298-n&i86957&i86957&09593427-n\\
aidoneus\%1:18:00::&09570298-n&i86957&i86957&09593427-n\\
hades\%1:18:00::&09570298-n&i86957&i86957&09593427-n\\
\cline{4-5}\noalign{\smallskip}
pluto\%1:18:00::&09570298-n&i86957&i86958&09593643-n\\
\cline{1-3}\noalign{\smallskip}
dis\%1:18:00::&09570522-n&i86958&i86958&09593643-n\\
orcus\%1:18:00::&09570522-n&i86958&i86958&09593643-n\\
dis\_pater\%1:18:00::&$\nexists$&$\nexists$&i86958&09593643-n\\
\hline
\end{tabular}
}
\end{center}

\medskip

The problem is that foreign language translations of the involved synsets
cannot deal with this change by simply applying a {\it concept to offset} mapping
like the Collaborative Interlingual index (CILI\footnote{\url{https://github.com/globalwordnet/cili}}).
In the French Wordnet, for example, {\it Pluton} is a synonym of {\it Hadès} and {\it Aides},
and thus a member of the Greek gods, and remains so, even after applying
the CILI mapping. Unlike the English {\it Pluto}, the French {\it Pluton} keeps the
CILI {\it i86957} identifier, and still translates to {\it Hades} in later English Wordnet versions.
Conversely, the French translation of the PWN 3.1 synset with CILI {\it i86958} does not
include {\it Pluton}. To adequately deal with this situation, the French
{\it Pluton} would need a link to the corresponding English sense key, instead of
being linked at the synset level.

Here, where both gods are the same and the name {\it Plouton} actually
exists in the Greek mythology, it would make sense to apply the {\it map-to-all}
strategy, and insert {\it Pluto} in both target synsets, as in the
mappings from the Sense Key Index (SKI)\footnote{\url{https://github.com/ekaf/ski}}.
But mapping to all possible targets is not guaranteed to be adequate in all cases,
so it is always preferable to review all the synset splits manually.

We aim to support wordnet interoperability in the general-purpose natural language toolkit
NLTK\footnote{\url{https://www.nltk.org}} \citep{bird09}, which is increasingly used in very diverse
Machine Learning projects, without specialized lexicographic knowledge. 
So a {\em one-to-many} synset mapping strategy would not be an
adequate default, because users would not know how to choose the most adequate target synset from a list
of mapping candidates. In such cases, it is more convenient that the system only picks one
target synset for each source synset.

Mapping the wordnets on the fly, at load time, requires an algorithm that performs as close to instantly
as possible, so we prefer a simple frequency-based approach, rather than a more complex analysis of
relation links. Therefore, we map each source synset to the target that retains most of the source
lemmas and, in the case of equality, to the synset with the highest offset. In most cases, though,
the choice is limited to one single target synset, since choosing between synsets is only relevant in
the cases where a source synset is split into two (or eventually three) synsets. These cases are rare
\cite{kafe18}, so candidates with an equal number of lemmas are even rarer.

So we apply a {\em many-to-one} mapping strategy, where potentially many (though most often only one) source
synsets are merged into a single target synset. This is the only difference between this work and the
{\em many-to-many} mappings from the Sense Key Index (SKI), 
resulting in slightly different numbers of False Positives
(fp) and False Negatives (fn), and only tiny differences in overall performance.

\subsection{Linear Time Algorithm}

\begin{algorithm*}
\caption{Map synsets from {\it source} to {\it target} Wordnet version using sense keys}
{\small
\label{alg1}
\begin{algorithmic}[0]
\State $\text{\sc sense\_index}_{source} \gets \left\{\,\forall\;\text{sense} \in \text{\it source}:
     \text{sense}_{key} \rightarrow \text{synset\_id}_{source}\,\right\}$
\State $\text{\sc sense\_index}_{target} \gets \left\{\;\forall\;\text{sense} \in \text{\it target}:
     \text{sense}_{key} \rightarrow \text{synset\_id}_{target}\;\right\}$
\State $\text{\sc map\_to\_many} \gets \left\{\,\forall\;\text{synset\_id}_{source} \in 
    \text{values}(\text{\sc sense\_index}_{source}):
    \text{synset\_id}_{source} \rightarrow \emptyset\,\right\}$
\For{ $\text{sense}_{key} \in \text{\sc sense\_index}_{source} \cap \text{\sc sense\_index}_{target}$}
    \State $\text{\sc map\_to\_many}[\text{synset\_id}_{source}].\text{append}(\text{synset\_id}_{target})$
\EndFor
\State $\text{\sc map\_to\_one} \gets \left\{\,\forall\;\text{synset\_id}_{source} \in \text{\sc map\_to\_many}:
    \text{synset\_id}_{source} \rightarrow \text{argmax}(\text{count}(\text{synset\_id}_{target}))\,\right\}$
\State {\bf return} $\text{\sc map\_to\_one}$
\end{algorithmic}
}
\end{algorithm*}

Algorithm \ref{alg1} constructs a mapping between two English Wordnet (PWN or OEWN) versions
(respectively $source$ and $target$), using intermediate mappings,
implemented here as Python dictionaries (see the NLTK listing in Appendix \ref{pylst}).

First, we construct a mapping from the sensekeys to the corresponding synset identifier
({\it synset\_id}) for each of the $source$ and $target$ Wordnet versions.
For this, we use either the {\it index.sense} file included in each PWN release,
or the {\it sense id} attribute of the OEWN senses, since OEWN now uses sensekeys directly
as its main sense identifier. NLTK does not yet support ILI identifiers, so the current
NLTK implementation can only use {\it offset-part\_of\_speech} synset identifiers, but it is
straightforward to replace these by ILI concept identifiers.
Each sensekey is linked to at most one synset in each version, but may
be absent from either the source or target version (in the cases where a sense was added or removed).
This step does one pass over the {\it index.sense}, which consists in one record per
sensekey, so its complexity is obviously linear.

Then the {\sc map\_to\_many} step joins the two {\sc index\_sense} maps in order to
produce a {\it synset\_to\_many} mapping from the {\it source} synset identifiers to lists of
corresponding synset identifiers in {\it target}. Python sets are implemented as hash tables,
with $O(1)$ lookup, so the intersection of both versions' sense keys is computed in $O(n)$ time.
Then we do one pass over the sources' offsets, to initialize empty candidate bags,
and one pass over the common sense keys, to populate the {\sc map\_to\_many} mapping,
which is identical to the corresponding SKI mapping \citep{kafe18}.

Finally, a {\sc map\_to\_one} step chooses the most adequate target synset
for each source synset, among a bag of candidates provided by the {\sc map\_to\_many} mapping.
This step is optional for use cases where we want to retain all the candidate targets.
Here, we use the $max$\footnote{Thanks to Steven Bird, who reviewed the initial implementation,
and pointed out that $max$ is quicker than $sort$.}
function to pick the target synset that retains most lemmas from the source synset, but
we also discuss using $sort$ as an alternative in section 4.3. We do one pass over each of the candidate bags,
where we use the $O(n)$ $max$ function to pick the target synset, so this step also runs in linear
time.

\subsection{Complexity}

Since each of its steps runs in linear time, the total complexity of this mapping algorithm
is also $O(n)$, where $n$ corresponds to the numbers of sense keys and synset offsets in the
involved wordnets. To our knowledge, this is the simplest
mapping algorithm yet proposed for wordnets, and considerably less complex than the deep relation
analysis in \citet{daude-etal-2000-mapping} and \citet{Daud01},
although both approaches have similar performance, but also complementary strengths and weaknesses \citep{kafe18}.

\subsection{Implementation}

We first integrated this mapping process in the {\em wordnet} library of NLTK version 3.6.6,
and used it to map the multilingual wordnets from OMW 1.4 \cite{bond-etal-2020-issues}
at load time, converting their PWN 3.0 synset identifiers
to those used in any of the more recent English Wordnets, in order to support the seamless interoperation
of the involved databases.

NLTK is developed on an open software development platform\footnote{\url{https://github.com/nltk/nltk}}, 
which provides free access for all, to not only the software code, but also its various incarnations,
and the corresponding discussions before and after its release.
Everyone is free to modify the source code, and welcome to contribute improvements
back to the community.

When using synset offsets, the implementation differs from algorithm \ref{alg1} by adding a
supplementary mapping link from adjectives, when the source synset is an adjective satellite.
This is necessary for handling OMW data, where most languages ignore the $satellite$ category.
But this step does not apply to ILI identifiers, since these don't include any
part-of-speech reference.

We rewrote the implementation for NLTK version 3.8, in order to closely follow algorithm \ref{alg1}.
In the initial implementation, the source wordnet was hard-coded to PWN version 3.0, for
handling the OMW data. An optional {\em version} parameter has been added in the forthcoming NLTK 3.8.2,
which allows to produce mappings for any pair of English Wordnet versions.
Appendix \ref{pylst} includes the listing of this slightly more elaborated implementation,
which additionally collects the {\em split} or {\em lost} synsets in structures called
respectively $splits$ and $nomap$, which should be useful for further improving the mappings.
We also adapted the functions in the appendix for the
Wn\footnote{\url{https://github.com/goodmami/wn}} library \cite{goodman-bond-2021-intrinsically},
in order to compare the performance of algorithm \ref{alg1} using respectively synset offsets
versus ILIs as synset identifiers. We thus used {\it Wn} to produce the $Map_{CILI}$ results in 
table \ref{tab:lgs}, while we computed the
$Map_{Offset}$ results in table \ref{tab:lgs} using both NLTK and {\it Wn}, and verified that both
libraries yield identical outputs.

\section{Results}

\subsection{Multilingual Coverage }

\begin{table*}
\begin{center}
\caption{\label{tab:lgs}Multilingual synsets in OMW 1.4 mapped to OEWN 2021 using synset offsets vs. CILI 1.0}

\resizebox{\textwidth}{!}{
\begin{tabular}[]{lrrrrrrr}
\hline
\noalign{\smallskip}
&{\bf Synsets}&\multicolumn{3}{c}{\bf Map$_{Offset}$}&\multicolumn{3}{c}{\bf Map$_{CILI}$}\\
\noalign{\medskip}
{\bf Language}&{\em PWN 3.0}&{\em OEWN 2021}&{\em Lost}&{\em \%}&{\em OEWN 2021}&{\em Lost}&{\em \%}\\
\noalign{\smallskip}
\hline
\noalign{\medskip}
{\em English}&117659&117454&{\bf205}&0.17&117427&232&0.20\\
{\em Finnish}&116763&116562&{\bf201}&0.17&116535&228&0.20\\
{\em Thai}&73350&73240&{\bf110}&0.15&73223&127&0.17\\
{\em French}&59091&59015&{\bf76}&0.13&59005&86&0.15\\
{\em Japanese}&57184&57086&{\bf98}&0.17&57080&104&0.18\\
{\em Romanian}&56026&55941&{\bf85}&0.15&55931&95&0.17\\
{\em Catalan}&45826&45773&{\bf53}&0.12&45769&57&0.12\\
{\em Portuguese}&43895&43844&{\bf51}&0.12&43840&55&0.13\\
{\em Slovenian}&42583&42520&{\bf63}&0.15&42513&70&0.16\\
{\em Mandarin Chinese}&42300&42249&{\bf51}&0.12&42240&60&0.14\\
{\em Spanish}&38512&38431&{\bf81}&0.21&38418&94&0.24\\
{\em Indonesian}&38085&38018&{\bf67}&0.18&38011&74&0.19\\
{\em Standard Malay}&36911&36843&{\bf68}&0.18&36836&75&0.20\\
{\em Italian}&35001&34964&{\bf37}&0.11&34960&41&0.12\\
{\em Polish}&33826&33798&{\bf28}&0.08&33794&32&0.09\\
{\em Dutch}&30177&30154&{\bf23}&0.08&30151&26&0.09\\
{\em Basque}&29413&29387&{\bf26}&0.09&29386&27&0.09\\
{\em Croatian}&23115&23081&{\bf34}&0.15&23077&38&0.16\\
{\em Galician}&19311&19290&{\bf21}&0.11&19283&28&0.14\\
{\em Slovak}&18507&18478&{\bf29}&0.16&18472&35&0.19\\
{\em Modern Greek (1453-)}&18049&18025&{\bf24}&0.13&18023&26&0.14\\
{\em Italian (iwn)}&15563&15553&{\bf10}&0.06&15553&{\bf10}&0.06\\
{\em Standard Arabic}&9916&9897&{\bf19}&0.19&9896&20&0.20\\
{\em Lithuanian}&9462&9446&{\bf16}&0.17&9442&20&0.21\\
{\em Swedish}&6796&6784&{\bf12}&0.18&6784&{\bf12}&0.18\\
{\em Hebrew}&5448&5441&{\bf7}&0.13&5439&9&0.17\\
{\em Bulgarian}&4959&4950&{\bf9}&0.18&4950&{\bf9}&0.18\\
{\em Icelandic}&4951&4942&{\bf9}&0.18&4942&{\bf9}&0.18\\
{\em Albanian}&4675&4668&{\bf7}&0.15&4668&{\bf7}&0.15\\
{\em Danish}&4476&4468&{\bf8}&0.18&4468&{\bf8}&0.18\\
{\em Norwegian Bokmål}&4455&4447&{\bf8}&0.18&4447&{\bf8}&0.18\\
{\em Norwegian Nynorsk}&3671&3666&{\bf5}&0.14&3666&{\bf5}&0.14\\
\noalign{\smallskip}
\hline
\noalign{\medskip}
{\em Average}&32811.12&32762.97&{\bf 48.16}&0.15&32757.16&53.97&0.16\\
\noalign{\smallskip}
\hline
\noalign{\medskip}

\end{tabular}
}
\end{center}
\caption*{We computed the Map$_{Offset}$ results using both the NLTK and {\it Wn}
software libraries, and the Map$_{CILI}$ results with only {\it Wn},
since NLTK does not yet support ILI identifiers.}
\end{table*}

Table \ref{tab:lgs} displays the number of synsets and lemmas in NLTK's data package for OMW 1.4,
when loaded with respectively the default PWN 3.0, and OEWN Edition 2021.
The languages are listed by their number of synsets in decreasing order, and we report
the number of synsets lost, as well as percentages, when mapping between the two
English Wordnet versions, using either synset offsets or the CILI 1.0 synset identifiers
currently included in the {\it Wn} library.

All the multilingual wordnets suffer a loss in the mapping, but this loss is almost
negligible with either type of synset identifier: at most $0.19\%$ (corresponding to
$99.81\%$ recall) for Standard Arabic with synset offsets, and $0.21\%$ using CILI
with Lithuanian. Except a small number of ties with the smallest wordnets, the synset
offset mappings perform better than the CILI 1.0 mappings in all cases.
This is surprising since the CILI mappings were partially curated manually, so we expected
them to provide an advantage over the completely automatic offset mappings.
However, the difference is small, and might be attributed to known issues
\footnote{CILI issue \#16, \url{https://github.com/globalwordnet/cili/issues/16}}
with the CILI 1.0 mappings, which could be remedied in a future version.

With PWN 3.0, some numbers are identical to those reported by \citet{Bond14}.
These concern wordnets that have not been updated since OMW 1.0. On the other hand, 
some wordnets in OMW 1.4
are not current, as for ex. the Basque, Catalan, Galician and Spanish wordnets date back
to the 2012 edition of the Multilingual Core Repository (MCR) described by \citet{Agir12},
although the coverage of these wordnets was greatly expanded in the 2016 edition of MCR.

NLTK also has a PWN 3.1 data package, where the mapping loss is usually less than half,
compared to OEWN 2021, and for ex. only $0.09\%$ for Standard Arabic, corresponding
to $99.91\%$ recall.
We also mapped two variants of OEWN Edition 2022: the official release
\footnote{\url{https://en-word.net/static/english-wordnet-2022.zip}},
and an alternative version provided by the
XEWN\footnote{\url{https://github.com/x-englishwordnet}} project.
Their databases have different sizes, and hence different synset offsets, but
both yielded identical mapping losses, which were slightly better than
OEWN 2021 in all cases, for ex. $0.17\%$ synset lost with Standard Arabic.
Standard mappings are not likely to become available for different variants of the
same Wordnet version, so an advantage of our method is that it nevertheless
allows a downstream comparison of these variants, which would not be possible
otherwise.

\subsection{Splits and Merges}
As a consequence of our mapping strategy, where we only pick one target for each source synset,
the synsets are never split. On the contrary, all lemmas belonging to a source synset, that
would be split according to a many-to-many strategy, are mapped to the same target synset, and synonymy
persists.

With the example from section 2.1, since {\it Pluto} is not split out of its {\it source} synset,
it is not {\it merged} into its {\it target} synset, but remains a synonym of the other Greek gods:
\smallskip
\begin{center}
\resizebox{0.48\textwidth}{!}{
\begin{tabular}[]{lrrrr}
{\bf Sense Key}&{\bf PWN}$_{3.0}$&{\bf CILI}$_{3.0}$&{\bf CILI}$_{3.1}$&{\bf PWN}$_{3.1}$\\
\hline\noalign{\smallskip}
aides\%1:18:00::&09570298-n&i86957&i86957&09593427-n\\
aidoneus\%1:18:00::&09570298-n&i86957&i86957&09593427-n\\
hades\%1:18:00::&09570298-n&i86957&i86957&09593427-n\\
pluto\%1:18:00::&09570298-n&i86957&i86957&09593427-n\\
\hline\noalign{\smallskip}
dis\%1:18:00::&09570522-n&i86958&i86958&09593643-n\\
orcus\%1:18:00::&09570522-n&i86958&i86958&09593643-n\\
dis\_pater\%1:18:00::&$\nexists$&$\nexists$&i86958&09593643-n\\
\hline
\end{tabular}
}
\end{center}

\medskip
The result is mostly a one-to-one mapping, with only 44 many-to-one cases occurring, when different
source synsets are merged into the same target synset.
Our method maps all the merged foreign language synsets
to their correct target, as for ex. with the {\it baseball} example below.
This contrasts with the current implementation of the {\it Wn} library's
standard {\it translate} function, which finds no translation for the first PWN$_{3.0}$ synset
({\it i37881}) in PWN$_{3.1}$. Conversely, translating {\em i37882} back from 
PWN$_{3.1}$ to PWN$_{3.0}$, {\em Wn} does not find the {\em i37881} lemmas.

\smallskip
\begin{center}
\resizebox{0.48\textwidth}{!}{
\begin{tabular}[]{lrrrr}
{\bf Sense Key}&{\bf PWN}$_{3.0}$&{\bf CILI}$_{3.0}$&{\bf CILI}$_{3.1}$&{\bf PWN}$_{3.1}$\\
\hline\noalign{\smallskip}
baseball\%1:04:00::&00471613-n:&i37881&i37882&00472688-n\\
baseball\_game\%1:04:00::&00471613-n:&i37881&i37882&00472688-n\\
\cline{1-3}\noalign{\smallskip}
ball\%1:04:01::&00474568-n&i37882&i37882&00472688-n\\
\hline
\end{tabular}
}
\end{center}

\medskip

The problem is that {\it Wn} only knows the correspondence between ILIs and offsets {\em within}
each involved Wordnet version, but has no mapping {\em between} these versions. Merged synsets
disappear in translation\footnote{\url{https://github.com/goodmami/wn/issues/179}}, 
because only one of the
merged CILI identifiers is available in the target, so the synsets with the other ILIs are no
longer reachable. This problem with merged ILIs in {\it Wn} only concerns a small number of synsets,
since each foreign language wordnet covers only a fraction
of the 44 merged English synsets. It does not affect the Map$_{CILI}$ results in
Table \ref{tab:lgs},
since we computed these using our mapping algorithm, instead of {\it Wn}'s standard
{\it translate} function.

\subsection{Performance}
We found that our method could not map 205 English synset offsets from PWN 3.0 to an OEWN 2021 target.
The small mapping losses in table \ref{tab:lgs} correspond to the subset of these 205 synsets included in
each multilingual wordnet. These losses represent all the {\em negatives} in
a confusion matrix, amounting to the addition of the {\em True Negatives} ({\it tn}),
which were truly removed in the target Wordnet, and the {\em False Negatives} ({\it fn}),
which we ideally should be able to map. So among the mapping losses, only the {\it fn} are fallacies.

The minority lemmas in the split English synsets, which are induly mapped to the same synset
as in the source, constitute the {\em False Positives} ({\it fp}). These only amount to the
44 splits between PWN 3.0 and OEWN 2021, so their number is small, compared
to the True Positives (117454 minus eventual sense key violations).


{\flushleft
\resizebox{0.50\textwidth}{!}{
\begin{tabular}[]{l@{\extracolsep{20pt}}ll}
{\it Synsets}&{\bf Mapped}&{\bf Not Mapped}\\
\hline\noalign{\smallskip}
{\bf True}&$PWN_{3.0} \cap OEWN_{2021}$&$\emptyset$\\
&$tp = 117454$&$tn = 0$\\
\noalign{\smallskip}
{\bf False}&$Splits$&$\complement_{OEWN_{2021}}^{PWN_{3.0}}$\\
&$fp = 44$&$fn = 205$\\
\end{tabular}
}
}

We evaluate the performance of our algorithm using the values above,
and obtain almost perfect performance results:
\begin{equation}
precision = \frac{tp}{tp+fp} = 0.9996
\end{equation}
\begin{equation}
recall = \frac{tp}{tp+fn} = 0.9983
\end{equation}
\begin{equation}
f1 = \frac{2*precision*recall}{precision+recall} = 0.9989
\end{equation}

Thus, the overall performance of the English mapping is $99.89\%$, which compares favorably with more
complex mapping strategies like \citet{daude-etal-2000-mapping}.

Comparing the lost English synsets between the two types of synset identifiers (offsets vs. ILIs),
we found that 143 were lost using both types,
while 62 were only lost with offsets (always due to satellite adjectives becoming
standard adjectives), and 89 were only lost with CILI 1.0. The respective additions of these losses yield
the total loss reported for English in table \ref{tab:lgs} (205 with offsets vs. 232 with the ILI).

\section{Discussion}

We have shown that mapping between different English Wordnet versions is feasible in linear time,
by relying on the stability of PWN sense keys.
Our method allows to transparently update the database links on-the-fly, to another
English Wordnet version, even though no prior mapping exists yet. This can benefit any
database linked with an English Wordnet, and enhance any downstream task that uses
such a database.

\subsection{Coverage and Integrity}
Our results show that almost all the vocabulary of
the multilingual wordnets in OMW 1.4 persisted after the mapping.

Some doubts remain necessarily, though, concerning the referential integrity
of the sensekeys, on which the mappings rely. Sensekeys are meant to always
refer to the same wordsense across wordnet versions, but \citet{kafe18}
reported a few violations of sensekeys' referential integrity. The number of these violations seems
negligible in PWN, but their impact has not yet been studied in OEWN. However, the fact that OEWN
now uses the PWN sensekeys as principal wordsense identifier, is a reason for considering
that the sensekeys are indeed persistent in OEWN, and that we can rely on their referential
integrity in theory. Still, it would be helpful to investigate in practice, whether the addition of a
new wordsense in OEWN could entail a modification of the sensekeys for other existing senses of the same word.

\subsection{Challenges and Opportunities}

In the mapping between PWN 3.0 and OEWN 2021, which we investigated here, our method displayed two
shortcomings: 205 English synsets were completely lost in the mapping, and 44 split synsets were
somewhat arbitrarily mapped to one single target. It is questionable, to which extent any
automatic mapping can provide linguistically satisfying targets for each of these cases.
Fortunately, their number is sufficiently small to allow a manual review, of which we can
already attempt to sketch some outlines.

It is possible, for ex., to identify genuinely lost synsets, which do not have any plausible target.
This happens when all the words included in the source synset are completely absent from
the target Wordnet version. Here, it occurred in particular with a number of racially tainted expressions,
like the synset $\{darky, darkie, darkey\}$, defined as "(ethnic slur) offensive term for Black people".
In these cases, relaxing the equivalence criteria, and mapping the synset to for ex. a superordinate,
would entail losing an essential nuance, and might often not be adequate. So these losses may be
unavoidable, unless choosing to retain the synset with its original meaning.

On the other hand, many losses are relatively easy to avoid. For example, out of the 205 English synsets
that our algorithm doesn't map, 62 concern adjective satellites which were changed to plain adjectives.
These have an obvious mapping through the ILI, where both Wordnet versions share the same concept identifier.

In other cases, we can identify changes in a part of the sense key,
for words that keep identical definitions. This reveals that unfortunate changes can occur in
any sense key part between two wordnet versions.
For example, Table \ref{tab:skchg} shows how the {\em lex\_id} of a sense of "sequoia" changed
from {\it 00} to {\it 01} between PWN 3.0 and OEWN 2021,
while the {\em lexfile} of a sense of "stub out" changed from {\it 30} to {\it 35},
the adjective category of "obtrusive" changed from {\it 3} to {\it 5},
and the satellites' head adjective of "newfangled" changed from {\it original} to {\it new}.

\begin{table}
\caption{\label{tab:skchg}Changed Sense Key Parts (Examples)}
\resizebox{0.48\textwidth}{!}{
\begin{tabular}[]{p{3.3cm}p{3.3cm}p{3.3cm}}
\hline
\noalign{\medskip}
{\bf Sense Key}&{\bf PWN 3.0}&{\bf OEWN 2021}\\
\noalign{\smallskip}
\hline
\noalign{\medskip}
sequoia\%1:20:00::&{\it either of two huge coniferous California trees that reach a height of 300 feet;
sometimes placed in the Taxodiaceae}&$\nexists$\\
sequoia\%1:20:01::&$\nexists$&{\it either of two huge coniferous California trees that reach a height of 300 feet;
sometimes placed in the Taxodiaceae}\\
\noalign{\smallskip}
\hline
\noalign{\medskip}
stub\_out\%2:30:00::&{\it extinguish by crushing}&$\nexists$\\
stub\_out\%2:35:01::&$\nexists$&{\it extinguish by crushing}\\
\noalign{\smallskip}
\hline
\noalign{\medskip}
obtrusive\%3:00:00::&{\it undesirably noticeable}&$\nexists$\\
obtrusive\%5:00:00\-:noticeable:00&$\nexists$&{\it undesirably noticeable}\\
\noalign{\smallskip}
\hline
\noalign{\medskip}
newfangled\%5:00:00\-:original:00&{\it (of a new kind or fashion) gratuitously new}&$\nexists$\\
newfangled\%5:00:00\-:new:00&$\nexists$&{\it (of a new kind or fashion) gratuitously new}\\
\noalign{\smallskip}
\hline
\noalign{\bigskip}
\end{tabular}

}
\end{table}
In all these cases, we see different sense keys pointing to the same word sense, and this is different
from {\em key violations} (one sense key pointing to different word senses). In some cases,
the English lexicographers could prevent this problem, but it can also be remedied downstream,
by an additional mapping link between the few changed sense keys, which would allow even
higher quality mappings. Our implementation (see Appendix \ref{pylst}) supports eventual
further improvements of the mappings through the {\it map\_to\_many} function, and by providing
the {\it splits} and {\it nomap} lists of problematic cases to study in greater depth.

\subsection{Variants of the mapping algorithm}

Applying our mapping algorithm to other synset identifiers than the offsets only requires a
simple modification of the initial $IndexSense$ function, while our two other functions remain
unchanged. So we extended our approach, to also map ILI concept identifiers instead of synset offsets.
This is not always practical yet though, because of inherent delays in the current attribution
process for new ILI identifiers\footnote{CILI issue \#9, \url{https://github.com/globalwordnet/cili/issues/9}}.

We applied $max$ to a list of candidate $(count, offset)$ pairs, in order to pick the target synset that retains
most lemmas from the source synset. As a consequence, in the case of equal counts, the $max$ function
picks the target synset with the highest offset. 
But instead of the highest offset, it would be
possible to use the $min$ function, and pick the lowest offset instead when the counts are equal.
Alternatively, this strategy can be implemented by taking the first pair in a
sorted list, eventually sorting the counts in decreasing order and the offsets in increasing order.
Generally, the lowest offset corresponds to a synset that was included in the PWN databases before
those with higher offsets, so the choice between using $min$ or $max$ often induces a preference
for older versus newer synsets. More research could be useful, in order to assess which difference
this choice makes in practice.

Concerning the complexity of {\it max}, which is $O(n)$ versus {\it sort},
which is $O(n.logn)$, their difference is not substantial here,
where $n$ represents the number of target $(count,offset)$ pairs,
which is normally one, and only two or three in the rare cases where the source synset is split.

\section{Conclusion}

We presented an algorithm for mapping wordnets, that runs in linear time,
thus moving the frontier of wordnet interoperability by allowing to almost
instantly combine different database versions, for which no prior mapping exists.
We illustrated this capability by combining the OMW with OEWN. Other potential
uses include seamlessly updating existing PWN links in any Wordnet-linked
semantic web database, to newer OEWN versions.

We saw how our mappings only lose tiny amounts of data when
mapping multilingual wordnets, which indicates that the performance of this approach
is comparable to the best results obtained with alternative strategies.

Now that OEWN has adopted the original PWN sensekeys as main wordense identifier,
we may expect that the proposed algorithm remains relevant with future OEWN versions.
However, if more wordnet resources start to use a common set of persistent identifiers
like the PWN sensekeys, mappings could become unnecesary between these resources,
as they would be natively interoperable.

\section*{Acknowledgments}
Thanks to Tom Aarsen and Steven Bird for their useful review of the NLTK implementation,
and to the anonymous GWC 2023 reviewers for their many detailed and accurate suggestions.
The final version of this article also benefited from helpful comments by participants at the
GWC 2023 presentation, in particular Francis Bond and Piek Vossen.

\bibliography{gwc23ek}

\begin{thebibliography}{18}
\expandafter\ifx\csname natexlab\endcsname\relax\def\natexlab#1{#1}\fi

\bibitem[{Bird et~al.(2009)Bird, Klein, and Loper}]{bird09}
Steven Bird, Ewan Klein, and Edward Loper. 2009.
\newblock \emph{Natural language processing with Python: analyzing text with
  the natural language toolkit}.
\newblock O'Reilly Media, Inc.

\bibitem[{Bond et~al.(2014)Bond, Fellbaum, Hsieh, Huang, Pease, and
  Vossen}]{Bond14}
Francis Bond, Christiane Fellbaum, Shu-Kai Hsieh, Chu-Ren Huang, Adam Pease,
  and Piek Vossen. 2014.
\newblock A multilingual lexico-semantic database and ontology.
\newblock In \emph{Towards the Multilingual Semantic Web}, pages 243--258.
  Springer.

\bibitem[{Bond et~al.(2020)Bond, Morgado~da Costa, Goodman, McCrae, and
  Lohk}]{bond-etal-2020-issues}
Francis Bond, Luis Morgado~da Costa, Michael~Wayne Goodman, John~Philip McCrae,
  and Ahti Lohk. 2020.
\newblock \href {https://aclanthology.org/2020.lrec-1.390} {Some issues with
  building a multilingual {W}ordnet}.
\newblock In \emph{Proceedings of the 12th Language Resources and Evaluation
  Conference}, pages 3189--3197, Marseille, France. European Language Resources
  Association.

\bibitem[{Bond et~al.(2016)Bond, Vossen, McCrae, and
  Fellbaum}]{bond-etal-2016-cili}
Francis Bond, Piek Vossen, John McCrae, and Christiane Fellbaum. 2016.
\newblock \href {https://aclanthology.org/2016.gwc-1.9} {{CILI}: the
  collaborative interlingual index}.
\newblock In \emph{Proceedings of the 8th Global WordNet Conference (GWC)},
  pages 50--57, Bucharest, Romania. Global Wordnet Association.

\bibitem[{Daud{\'e} et~al.(2000)Daud{\'e}, Padr{\'o}, and
  Rigau}]{daude-etal-2000-mapping}
J.~Daud{\'e}, L.~Padr{\'o}, and G.~Rigau. 2000.
\newblock \href {https://doi.org/10.3115/1075218.1075282} {Mapping {W}ord{N}ets
  using structural information}.
\newblock In \emph{Proceedings of the 38th Annual Meeting of the Association
  for Computational Linguistics}, pages 504--511, Hong Kong. Association for
  Computational Linguistics.

\bibitem[{Daud\'{e} et~al.(2001)Daud\'{e}, Padr\'{o}, and Rigau}]{Daud01}
J.~Daud\'{e}, L.~Padr\'{o}, and G.~Rigau. 2001.
\newblock A complete wn1.5 to wn1.6 mapping.
\newblock In \emph{Proceedings of the NAACL Workshop 'WordNet and Other Lexical
  Resources: Applications, Extensions and Customizations' (NAACL'2001).,
  Pittsburg, PA, USA}.

\bibitem[{Fellbaum(1998)}]{WordNet}
Christiane Fellbaum. 1998.
\newblock \emph{WordNet, An Electronic Lexical Database}.
\newblock MIT Press, Cambridge.

\bibitem[{Gonzalez-Agirre et~al.(2012)Gonzalez-Agirre, Laparra, and
  Rigau}]{Agir12}
A.~Gonzalez-Agirre, E.~Laparra, and G.~Rigau. 2012.
\newblock Multilingual central repository version 3.0: upgrading a very large
  lexical knowledge base.
\newblock In \emph{Proceedings of the Sixth International Global WordNet
  Conference (GWC2012). Matsue, Japan}.

\bibitem[{Goodman and Bond(2021)}]{goodman-bond-2021-intrinsically}
Michael~Wayne Goodman and Francis Bond. 2021.
\newblock \href {https://aclanthology.org/2021.gwc-1.12} {Intrinsically
  interlingual: The wn python library for wordnets}.
\newblock In \emph{Proceedings of the 11th Global Wordnet Conference}, pages
  100--107, University of South Africa (UNISA). Global Wordnet Association.

\bibitem[{Kafe(2018)}]{kafe18}
Eric Kafe. 2018.
\newblock \href {https://doi.org/10.11649/cs.1717} {Persistent semantic
  identity in wordnet}.
\newblock \emph{Cognitive Studies | Études cognitives}, 18.

\bibitem[{McCrae et~al.(2020)McCrae, Rademaker, Rudnicka, and
  Bond}]{mccrae-etal-2020-english}
John~Philip McCrae, Alexandre Rademaker, Ewa Rudnicka, and Francis Bond. 2020.
\newblock \href {https://aclanthology.org/2020.mmw-1.3} {{E}nglish {W}ord{N}et
  2020: Improving and extending a {W}ord{N}et for {E}nglish using an
  open-source methodology}.
\newblock In \emph{Proceedings of the LREC 2020 Workshop on Multimodal Wordnets
  (MMW2020)}, pages 14--19, Marseille, France. The European Language Resources
  Association (ELRA).

\bibitem[{Navigli and Ponzetto(2010)}]{Babel}
Roberto Navigli and Simone~Paolo Ponzetto. 2010.
\newblock Babelnet: Building a very large multilingual semantic network.
\newblock In \emph{Proceedings of the 48th Annual Meeting of the Association
  for Computational Linguistics, Uppsala, Sweden, 11-16 July 2010}, pages
  216--225.

\bibitem[{Nielsen(2018)}]{Imagenet}
Finn~Årup Nielsen. 2018.
\newblock Linking imagenet wordnet synsets with wikidata.
\newblock In \emph{Proceedings of The 2018 Web Conference Companion (WWW'18
  Companion)}. ACM, New York, USA.

\bibitem[{Niles and Pease(2003)}]{Sumo}
Ian Niles and Adam Pease. 2003.
\newblock Linking lexicons and ontologies: Mapping wordnet to the suggested
  upper merged ontology.
\newblock In \emph{Ike}, pages 412--416.

\bibitem[{Suchanek et~al.(2008)Suchanek, Kasneci, and Weikum}]{Yago}
Fabian~M Suchanek, Gjergji Kasneci, and Gerhard Weikum. 2008.
\newblock Yago: A large ontology from wikipedia and wordnet.
\newblock \emph{Web Semantics: Science, Services and Agents on the World Wide
  Web}, 6(3):203--217.

\bibitem[{Vossen(2002)}]{Voss02}
Piek Vossen. 2002.
\newblock \emph{EuroWordnet General Document}.
\newblock EWN.

\bibitem[{Vossen et~al.(2016)Vossen, Bond, and
  McCrae}]{vossen-etal-2016-toward}
Piek Vossen, Francis Bond, and John McCrae. 2016.
\newblock \href {https://aclanthology.org/2016.gwc-1.59} {Toward a truly
  multilingual {G}lobal{W}ordnet grid}.
\newblock In \emph{Proceedings of the 8th Global WordNet Conference (GWC)},
  pages 424--431, Bucharest, Romania. Global Wordnet Association.

\bibitem[{WordNet-team(2010)}]{wnm}
WordNet-team. 2010.
\newblock Wordnet 3.0 reference manual.
\newblock In \emph{WordNet Documentation}. Princeton University,
  \url{https://wordnet.princeton.edu/documentation}.

\end{thebibliography}
\onecolumn
\appendix
\section{Appendix: Implementation in NLTK (Python)}
\label{pylst}
\resizebox{\textwidth}{!}{
\begin{lstlisting}[language=Python,firstline=1221,lastline=1284]
    def index_sense(self, version=None):
        """Read sense key to synset id mapping from index.sense file in corpus directory"""
        fn = "index.sense"
        if version:
            from nltk.corpus import CorpusReader, LazyCorpusLoader

            ixreader = LazyCorpusLoader(version, CorpusReader, r".*/" + fn)
        else:
            ixreader = self
        with ixreader.open(fn) as fp:
            sensekey_map = {}
            for line in fp:
                fields = line.strip().split()
                sensekey = fields[0]
                pos = self._pos_names[int(sensekey.split("%")[1].split(":")[0])]
                sensekey_map[sensekey] = f"{fields[1]}-{pos}"
        return sensekey_map

    def map_to_many(self, version="wordnet"):
        sensekey_map1 = self.index_sense(version)
        sensekey_map2 = self.index_sense()
        synset_to_many = {}
        for synsetid in set(sensekey_map1.values()):
            synset_to_many[synsetid] = []
        for sensekey in set(sensekey_map1.keys()).intersection(
            set(sensekey_map2.keys())
        ):
            source = sensekey_map1[sensekey]
            target = sensekey_map2[sensekey]
            synset_to_many[source].append(target)
        return synset_to_many

    def map_to_one(self, version="wordnet"):
        self.nomap[version] = set()
        self.splits[version] = {}
        synset_to_many = self.map_to_many(version)
        synset_to_one = {}
        for source in synset_to_many:
            candidates_bag = synset_to_many[source]
            if candidates_bag:
                candidates_set = set(candidates_bag)
                if len(candidates_set) == 1:
                    target = candidates_bag[0]
                else:
                    counts = []
                    for candidate in candidates_set:
                        counts.append((candidates_bag.count(candidate), candidate))
                    self.splits[version][source] = counts
                    target = max(counts)[1]
                synset_to_one[source] = target
                if source[-1] == "s":
                    # Add a mapping from "a" to target for applications like omw,
                    # where only Lithuanian and Slovak use the "s" ss_type.
                    synset_to_one[f"{source[:-1]}a"] = target
            else:
                self.nomap[version].add(source)
        return synset_to_one

    def map_wn(self, version="wordnet"):
        """Mapping from Wordnet 'version' to currently loaded Wordnet version"""
        if self.get_version() == version:
            return None
        else:
            return self.map_to_one(version)
\end{lstlisting}
}

\end{document}